\documentclass{bmvc2k}

\title{Transfer Learning by Ranking for Weakly Supervised Object Annotation}

\addauthor{Zhiyuan Shi}{zhiyuan.shi@eecs.qmul.ac.uk}{1}
\addauthor{Parthipan Siva}{psiva@eecs.qmul.ac.uk}{1}
\addauthor{Tao Xiang}{txiang@eecs.qmul.ac.uk}{1}

\addinstitution{
 School of Electronic Engineering and Computer Science,\\
 Queen Mary, University of London,\\
 London E1 4NS, UK
}

\runninghead{Shi,Siva,Xiang}{Transfer Learning by Ranking}

\def\etal{\emph{et al}\bmvaOneDot}
\def\voc{{\sc voc}}
\def\vocs{{\sc voc} 2007}
\def\pascal07{{\sc pascal voc} 2007}
\def\mil{{\sc mil}}
\def\misvm{{\sc mi-svm}}
\def\bow{{{\sc b}o{\sc w} }}
\begin{document}

\maketitle

\begin{abstract}
Most existing approaches to training object detectors rely on fully supervised learning, which requires the tedious manual annotation of object location in a training set. Recently there has been an increasing interest in developing weakly supervised approach to detector training where the object location is not manually annotated but automatically determined based on binary (weak) labels indicating if a training image contains the object. This is a challenging problem because each image can contain many candidate object locations which partially overlaps the object of interest. Existing approaches focus on how to best utilise the binary labels for object location annotation. In this paper we propose to solve this problem from a very different perspective by casting it as a transfer learning problem. Specifically, we formulate a novel  transfer learning based on learning to rank, which effectively transfers a model for automatic annotation of object location from an auxiliary dataset to a target dataset with completely unrelated object categories. We show that our approach outperforms existing state-of-the-art weakly supervised approach to annotating objects in the challenging \voc~dataset.
\end{abstract}


\section{Introduction}
\label{sec:intro}
\begin{figure}
\begin{minipage}[b]{4cm}
\subfigure[]{
\fbox{\includegraphics[width=3.9cm,height=4cm]{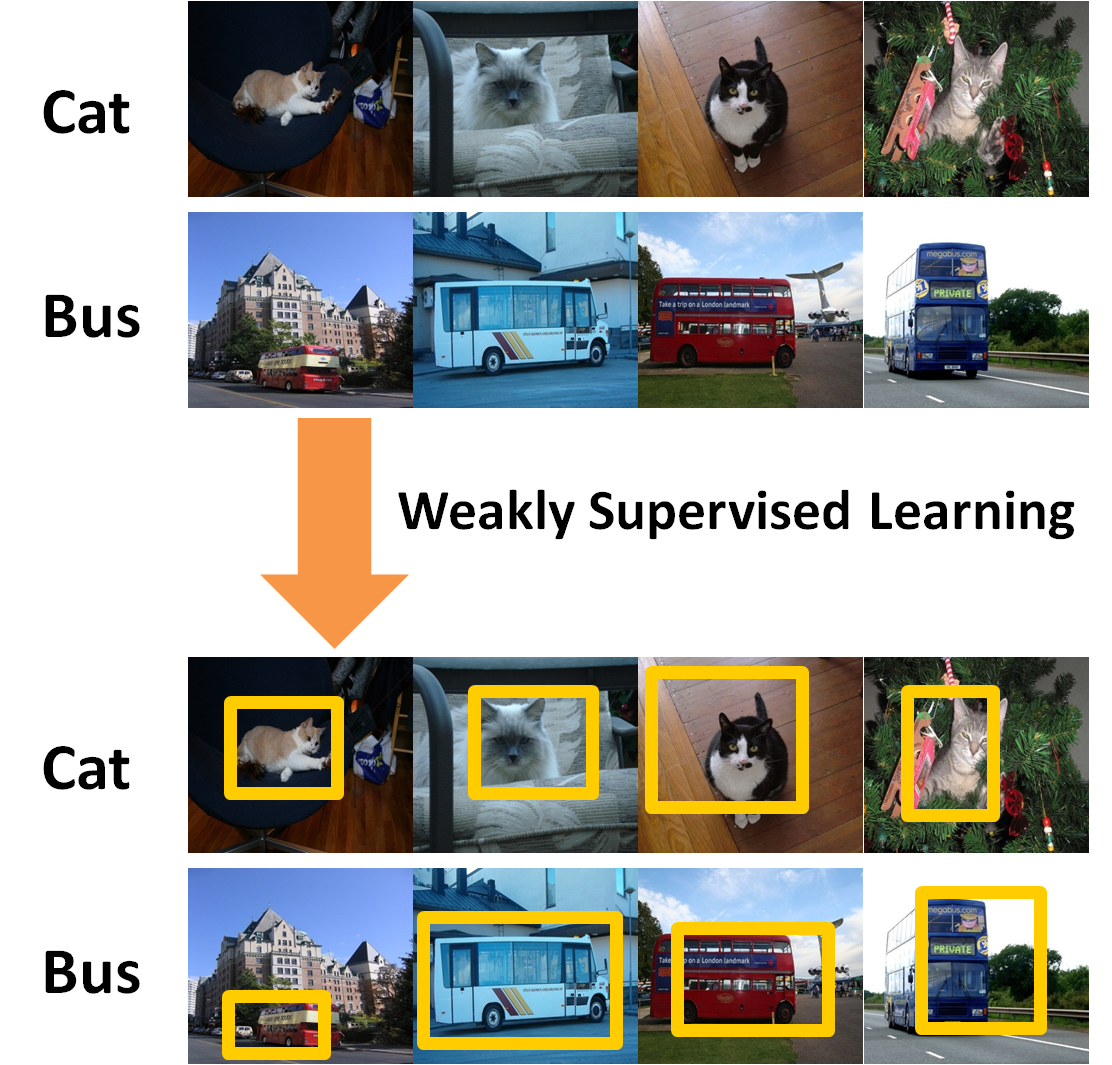}}
\label{1a}
}
\end{minipage}
\hspace{0.1cm}
\begin{minipage}[b]{4cm}
\subfigure[]{
\fbox{\includegraphics[width=3.9cm,height=4cm]{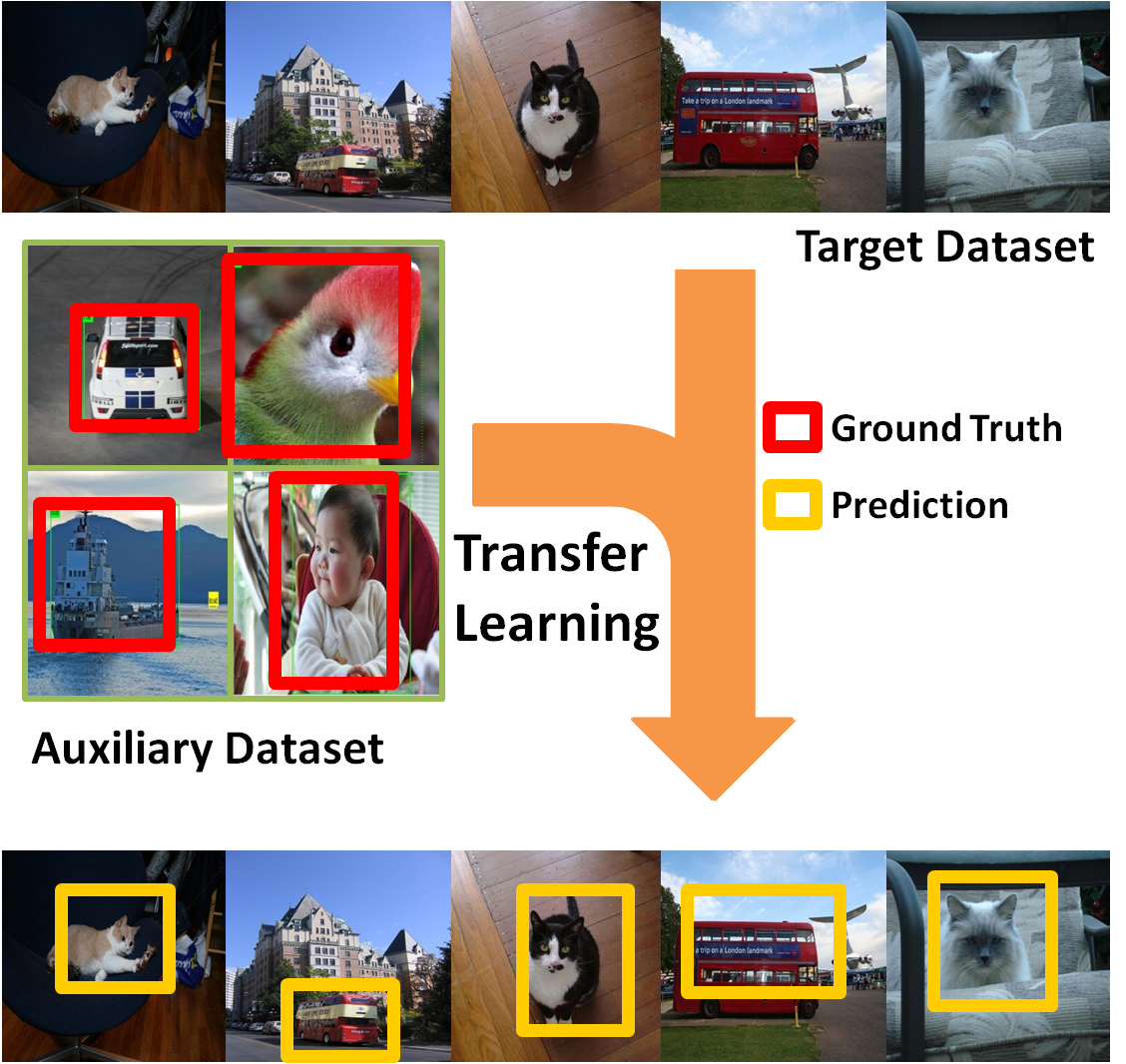}}
\label{1b}
}
\end{minipage}
\hspace{0.1cm}
\begin{minipage}[b]{4cm}
\subfigure[]{
\fbox{\includegraphics[width=3.9cm,height=4cm]{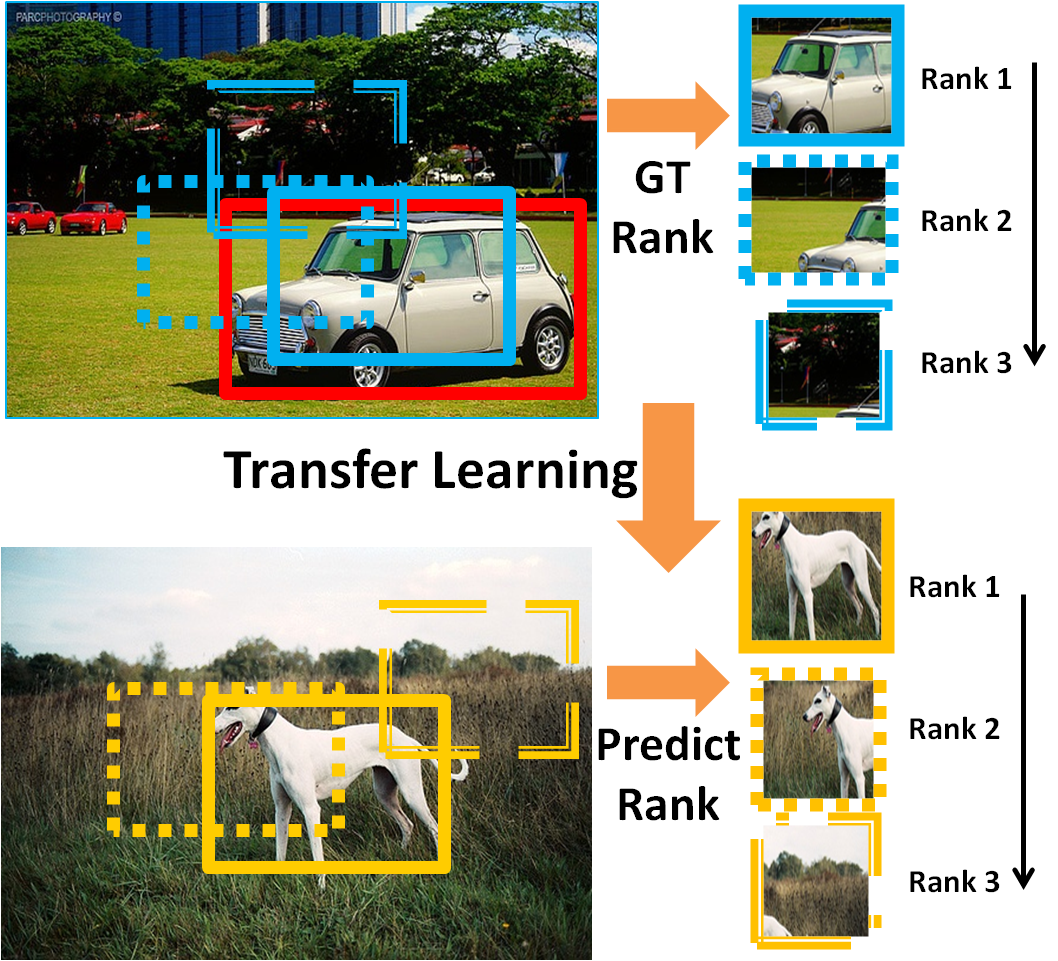}}
\label{1c}
}
\end{minipage}

\caption{  (a) Weakly supervised learning approach for automatic annotation of objects \cite{Deselaerslocalizing2010,Sivaiccv2011}. (b) Our transfer learning approach for automatic annotation of objects. (c) The mapping relationship between the degree of overlap and the appearance similarity between salient regions and the ground truth location is transferred from the auxiliary data to the target data.}
\label{fig:knowledge}
\end{figure}

Object detectors~\cite{Viola2001rapid,Felzenszwalb2012partbased} locate objects of interest in images and have many applications including image tagging, consumer photography, and surveillance. Most existing object detectors take a fully supervised learning (FSL) approach, where all the training images are manually annotated with the object location. However, manual annotation of hundreds of object categories is time-consuming, laborious, and subjective to human bias. To reduce the amount of manual annotation, a weakly supervised learning (WSL) \cite{Deselaerslocalizing2010, Sivaiccv2011, Nguyenweakly2011, Pandeyiccv2011} approach is desired. In WSL, the training set is only annotated with a binary label indicating the presence or absence of the object of interest, not the location or extent of the object (Fig.~\ref{1a}). WSL approaches first locate the object of interest in the training images and then the location information is used to train a detector in a fully supervised fashion.

Typically three information cues, saliency, inter-class, and intra-class, are used to locate or annotate the object of interest in images known to contain the object of interest (positive images). Saliency information ensures that the annotated region is a foreground region. Inter-class information ensures that the annotated regions look dissimilar to all images without the object of interest (negative images). Intra-class information ensures that the annotated regions in all positive images look similar to each other. Methods that use saliency alone \citep{Alexewhatisobject} select salient regions in each positive image independently. Methods that use inter-class and intra-class information \cite{Deselaerslocalizing2010,Sivaiccv2011} typically use saliency to limit the search space of each image by only looking at the most salient regions; then they select one of these salient regions by maximising the inter-class and intra-class information.

In this paper we utilise a fourth information cue (Fig.~\ref{1b}) which is typically neglected by other approaches: an auxiliary fully annotated dataset. While we want to reduce manual annotation when learning new object categories, we cannot ignore the fact that there exist many datasets which already have manual annotation of object locations \cite{pascalvoc2007}. However, these auxiliary datasets seem unhelpful since they often contain object categories that are unrelated to the target object category we wish to annotate. For example, an auxiliary dataset might contain annotations of cars, birds, boats and person but a target object category might be cats and buses (Fig.~\ref{1b}). So what information can we actually transfer? When adopting the strategy of selecting the optimal object location from a set of candidate salient regions \cite{Deselaerslocalizing2010,Sivaiccv2011}, the performance of the selection can obviously be measured by examining the degree of overlap between the selected region and the ground truth region (Fig.~\ref{1c}). One can safely assumes that the more a salient region overlaps the ground truth region, the more similar the two's appearances are. In other words, there exists a mapping relationship between the degree of overlap (hence the accuracy of annotation) and the appearance similarity. This relationship should hold true regardless of the object category  and is what we propose to learn and transfer to the target data. To quantify this mapping relationship, one must take into consideration the high dimensionality typical for representing object appearance and the inevitable noise. To this end, we formulate a ranking based transfer learning model which, once learned,  takes appearance similarity as input and predicts the ranking order among all the candidate salient regions according to their degree of overlap with the (unknown) true object location.   We show that our novel transfer learning model outperforms the state-of-the-art WSL approaches on the challenging \pascal07 dataset. 


\noindent \textbf{Related Work.} Early works on weakly supervised annotation mainly focused on saliency based approaches \cite{Alexewhatisobject,Chum07,Russell06}. While these methods provided a set of potential salient object location, they were shown to perform poorly for automatic annotation of objects in challenging cluttered images \cite{Deselaerslocalizing2010}. Recently many methods \cite{Nguyenweakly2011,Deselaerslocalizing2010,Sivaiccv2011} re-cast the weakly supervised problem as a multiple-instance learning (\mil) problem. In a \mil~formulation, each image with the object of interest is treated as a positive bag with many instances (potential object locations) of which at least one is positive and the images without the object of interest are treated as negative bags with only negative instances. The \mil~based algorithms iteratively select the positive instances in each positive bag using inter-class and/or intra-class information. The approach by Nguyen \etal \cite{Nguyenweakly2011} is an inter-class method that defines the entire positive image as the initial positive instance and then trains a support vector machine (SVM) to separate these initial positive instances and the negative images. The trained SVM is then used as a detector on the positive training images to refine the object location. However, the initial assumption is that the entire image is a good representation of the object, which is not always true. Pandey and Lazenbnilk \cite{Pandeyiccv2011} relaxed this assumption by using a latent SVM that treats the actual location of the object as a latent variable which is a constraint to be at least 40\% overlapped with the entire image. Unlike the inter-class information based methods \cite{Nguyenweakly2011,Pandeyiccv2011}, Deselaers \etal \cite{Deselaerslocalizing2010} and Siva and Xiang \cite{Sivaiccv2011} use saliency \cite{Alexewhatisobject} to define the initial instances in each image. Then the positive instances are iteratively selected by optimising a cost function based on both inter-class and intra-class information. We show that our transfer learning approach using an auxiliary dataset can outperform annotation accuracy of these \mil~methods without using either intra or inter-class information from target data.

Most existing transfer learning methods in computer vision address the classification problem not the WSL annotation problem. They typically require the target and auxiliary classes to be from either the same class but cross-domains \cite{Pan_2008,Yangcrossdomain,Sinnotransfercomponent} (such as news video from different countries) or different but related classes \cite{zweig07_iccv} (such as giraffe and horse). In comparison, our model does not assume that the auxiliary data and target data are related; it is thus a more generally applicable method. There have been a couple of recent efforts on transferring knowledge between unrelated categories \cite{Zhengtransfer2011,Rainaselftaought07}, but they focus on image categorisation where each image is dominated by a single object. Notice that Deselaers \etal \cite{Deselaerslocalizing2010} also uses an auxiliary dataset for weakly supervised annotation. However, in their work auxiliary data is used for parameters tuning rather than learning a transferrable model. In contrast, we use a ranking model to learn and transfer knowledge between unrelated categories, which to our knowledge has never been attempted before.

\section{Proposed Approach}
\label{sec:approach}

In our approach we have an auxiliary dataset, $\mathcal{A}$, with a set of fully annotated images (Fig.~\ref{1b}); each image contains an object with its location manually annotated by a bounding box. Given a target image $T$, containing an object not in $\mathcal{A}$, our goal is to annotate the object location in $T$. As in \cite{Deselaerslocalizing2010,Sivaiccv2011}, for all images in the auxiliary dataset $\mathcal{A}$ and the target image $T$ we select the top $N$ salient regions returned by the generic object detector proposed in \cite{Alexewhatisobject} as potential object locations. Then our goal is to select one of the $N$ salient regions in the target image $T$ as the object annotation. We can think of selecting the best salient region as a ranking problem. We wish to learn a model from $\mathcal{A}$ and use it to rank the $N$ salient regions in $T$ such that the highest ranked region has the greatest overlap with the unknown true location of the object (Fig.~\ref{1c}).
 
\subsection{Feature}
\label{sec:Tansfer_Feature}

Given that the object in the target image $T$ is unrelated to the objects in the auxiliary dataset $\mathcal{A}$ we need to develop a feature that is independent of object category which can still be used to learn the ranking of salient regions based on the degree of ground truth overlap. To this end, we extract appearance features and compute the absolute feature difference between candidate regions and ground truth as input for learning a ranking model. 

More specifically, for each image $i \in \mathcal{A}$, we represent each of the $N$ salient regions with an unnormalised \bow histogram $x_{i,j}$, where $j=1 \ldots N$. To compute a feature from $x_{i,j}$ that is independent of the object category we define a difference vector $d_{i,j}$ as the feature of interest:
\begin{equation}
d_{i,j} = \left| \frac{x_{i,j}}{\| x_{i,j} \|_1} - \frac{g_{i}}{\| g_{i} \|_1 } \right|,
\label{eq:diffgt}
\end{equation}
where $|\cdot|$ indicates the element wise absolute difference, $\| \cdot \|_1 $ is the L1 norm and $g_{i}$ is the unnormalised \bow histogram of the manually annotated ground truth region in image $i$. However, by this definition the target image $T$ has a difference vector:
\begin{equation}
d^T_{j} = \left| \frac{x^T_{j}}{\| x^T_{j} \|_1} - \frac{g^T}{\| g^T \|_1 } \right|,
\label{eq:diffT}
\end{equation} 
which requires us to know $g^T$, the ground truth \bow histogram. Since the true location of the object in $T$ is exactly what we are after, we do not know $g^T$ in the target class images. To overcome this problem we approximate the ground truth \bow histogram in Eq. \ref{eq:diffT} by the average \bow histogram of the $N$ salient regions in the target image $T$:

\begin{equation}
g^T \approx \mu^T = \frac{1}{N} \sum_{j=1}^N x^T_{j}.
\label{eq:tapp}
\end{equation}

\noindent The resulting feature vectors is

\begin{equation}
\hat{d}^T_{j} = \left| \frac{x^T_{j}}{\| x^T_{j} \|_1} - \frac{\mu^T_{i}}{\| \mu^T_{i} \|_1 } \right|.
\label{eq:diffTapp}
\end{equation}

However, now there is a discrepancy between training, where we directly use the ground truth \bow histogram, and testing, where we approximate the ground truth \bow histogram. To unify the training and testing process, we also approximate the ground truth \bow histogram in the training process for computing $d_{i,j}$. That is for the auxiliary dataset we replace Eq. \ref{eq:diffgt} with: 


\noindent\begin{minipage}{0.5\linewidth}
\begin{equation}
g_{i} \approx \mu_{i} = \frac{1}{N} \sum_{j=1}^N x_{i,j}
\label{eq:auxapp}
\end{equation}
\end{minipage}
\begin{minipage}{0.5\linewidth}
\begin{equation}
\hat{d}_{i,j} = \left| \frac{x_{i,j}}{\| x_{i,j} \|_1} - \frac{\mu_{i}}{\| \mu_{i} \|_1 } \right|
\label{eq:diffijapp}
\end{equation}
\end{minipage}

\noindent Note that the ground truth annotation of the auxiliary dataset is still needed as it is used in Section \ref{sec:Ranking_SVM} to determine the ranking order of $\hat{d}_{i,j}$. In Section \ref{sec:experiments} we conduct experiments to show that the approximation of the ground truth \bow histogram (Eq. \ref{eq:diffijapp}) is a better strategy.


\subsection{Ranking Model}
\label{sec:Ranking_SVM}

\begin{figure}
\begin{center}
\fbox{\includegraphics[width=7cm]{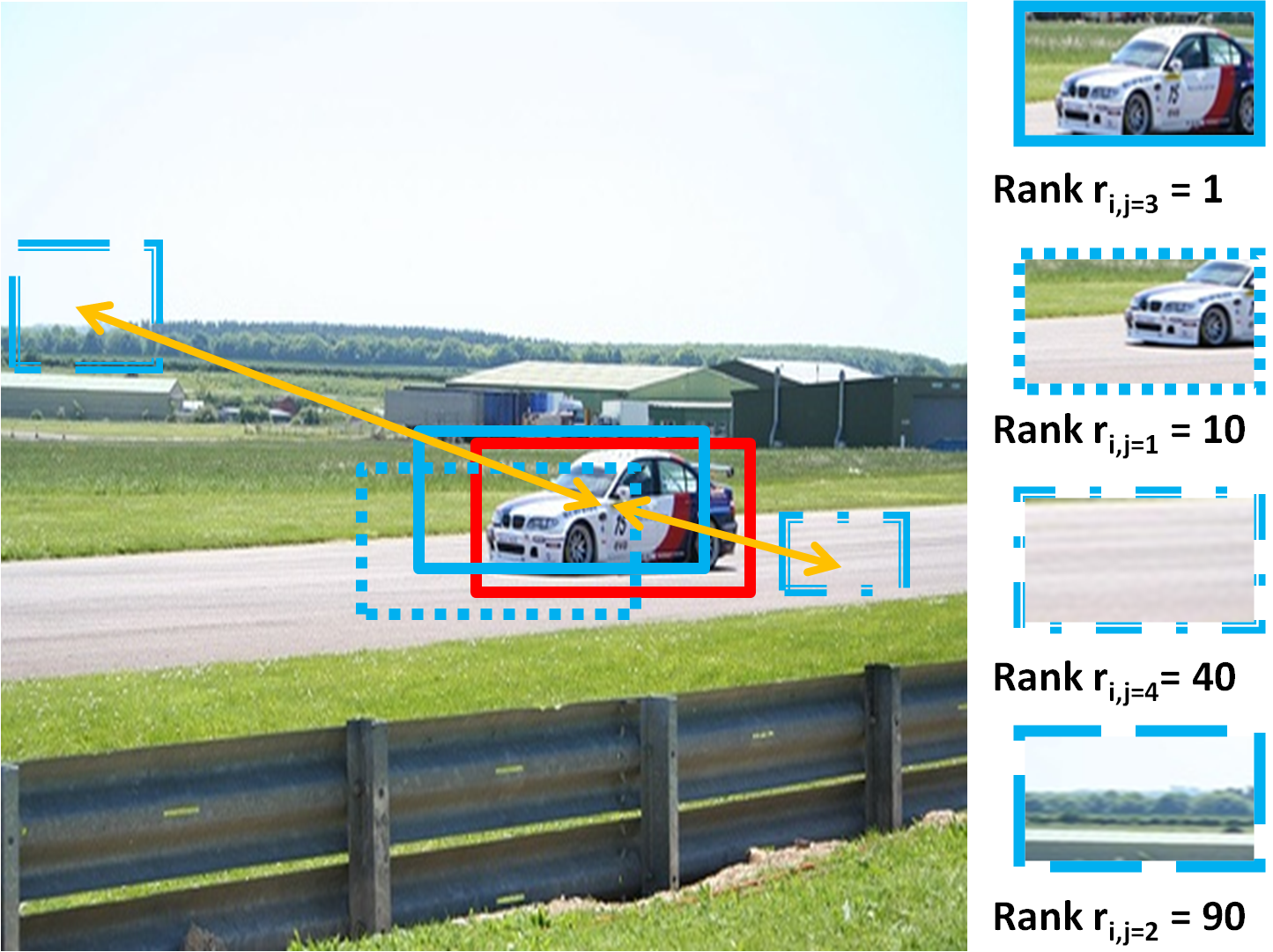}}
\end{center}

\caption{A higher rank is given to salient regions (blue) with higher overlap to the ground truth (red). In this image, a road region ($j=4$) which contains more relevant context to the car, is ranked higher than a sky region ($j=2$) according to their distances to the ground truth region (red).}
\label{fig:transfer_learning}
\end{figure}

For each image $i \in \mathcal{A}$ we now have $N$ salient regions represented by its corresponding difference vector $\hat{d}_{i,j}$, where $j=1 \ldots N$.  All $N$ salient regions are sorted by its overlap with the ground truth bounding box, where overlap is defined by \cite{pascalvoc2007} as the intersect area divided by union area. The sorted index of the salient region $j$ is used as the rank $r_{i,j}$ of the difference vector $\hat{d}_{i,j}$, where the salient region with the greatest overlap with the ground truth bounding box is given a rank of $1$. Salient regions with no overlap are ranked by their distance to the ground truth bounding box because regions nearer the ground truth location is likely to contain more relevant contextual information compared to regions that are farther away (Fig.~\ref{fig:transfer_learning}).

The difference vector and rank order pairs $\{(\hat{d}_{i,j},r_{i,j}) | i \in \mathcal{A}, j = 1 \ldots N \}$ are used to train a RankSVM \cite{Joachims_2002}. RankSVM is an ideal choice because it is able to cope with high dimensional feature spaces and large scale learning, exactly the problems faced by learning an object annotation ranking model.
RankSVM was originally introduced to improve the performance of the Internet search engine. The text document search/retrieval problem is similar to our problem in that it ranks the search results for a query based on how relevant the search result is to the query. In our case we wish to rank the salient regions based on how similar (relevant) it is to the ground truth in terms of visual appearance. In this context we have a set of queries $i$ (images $i \in \mathcal{A}$) and for all images we have a set of preference pairs $\left\lbrace (k,l) \in \mathcal{P} | r_{i,k} < r_{i,l} \forall k,l = 1 \ldots N, \forall i \in \mathcal{A} \right\rbrace$, indicating the ranking relationships between each pair of salient regions in each image. The number of preference pairs will be enormous for a moderate number of images and salient regions in each image ($N$). For efficient learning, we use the primal-based pairwise RankSVM algorithm proposed in \cite{Chapelle_2010} to minimise the objective function:

\begin{equation}
\frac{1}{2} \| \mathbf{w} \|^2 + C \sum_{(k,l) \in \mathcal{P}} \ell(\mathbf{w}^T \hat{d}_{i,k} - \mathbf{w}^T \hat{d}_{i,l}),
\end{equation}

\noindent where $\ell(t) = max(0,1-t)^2$ is the loss function and the weight $C$ is obtained by cross-validation.

\subsection{Annotating Target Image}

Given a target image $T$ we first obtain the $N$ salient regions using \citep{Alexewhatisobject}. For each salient region, we obtain the difference vector $\hat{d}^T_j$ per Eq.~\ref{eq:diffTapp}. All difference vectors $\hat{d}^T_{j=1\ldots N}$ are then ranked by the RankSVM trained on the auxiliary dataset. The salient region with the highest rank is then selected as the annotation for the target image $T$. Note that we are not using the weak annotation information, regarding if the target image $T$ contains the object of interest or not.

\section{Experiments}
\label{sec:experiments}

\noindent \textbf{Dataset --} All experiments are conducted on the challenging \vocs~dataset \cite{pascalvoc2007}. We use the \textit{AllView} dataset defined in \cite{Sivaiccv2011} for comparison and it consists of all 20 classes from the \vocs~training and validation set with no pose annotation. For our transfer learning model, we randomly choose 10 classes as the auxiliary dataset. After training the ranking model on these 10 classes, we annotate the remaining 10 target classes by our model.  The random selection is repeated 10 times and we report the average result on all 20 classes. As in \cite{Pandeyiccv2011,Sivaiccv2011}, for all 20 classes we exclude images annotated as difficult. We consider the predicted annotation (bounding box) as correct if it has more than 50\% overlap with ground truth bounding box as defined in \cite{pascalvoc2007} and report the percentage of correctly annotated images for performance measurement.

\noindent \textbf{Features --} Regular grid SIFT descriptors are computed on all images in the training data and quantized into 2000 words using k-means clustering. Each of the $N=100$ salient bounding boxes in an image, obtained by \cite{Alexewhatisobject}, is then described using a \bow histogram of 2000 bins.

\begin{table}[h]
\begin{center}
\begin{tabular}{l r @{.} l}
\hline
\textbf{Method} & 
\multicolumn{2}{c}{ \textbf{\emph{AllView} 20 Class}} \\
\hline
Ranking Model & \quad \quad \quad \quad \textbf{32}&\textbf{13\%} \\
\hline
Siva and Xiang \cite{Sivaiccv2011} & 29&29\% $^{*}$ \\
\hline
MI-SVM \cite{Andrews03supportvector} & 25&12\%\\
\hline
Objectness \cite{Alexewhatisobject} & 24&12\%\\
\hline
\multicolumn{3}{l}{{\scriptsize$^{*}$This number is computed based on the per class accuracy reported in \cite{Sivaiccv2011}.}}
\end{tabular}

\end{center}

\caption{Annotation results for \emph{AllView} dataset}
\label{results_ranking}
\end{table}

\begin{figure}[ht]
\begin{minipage}[b]{6cm}
\centering
\fbox{\includegraphics[height=6.6cm]{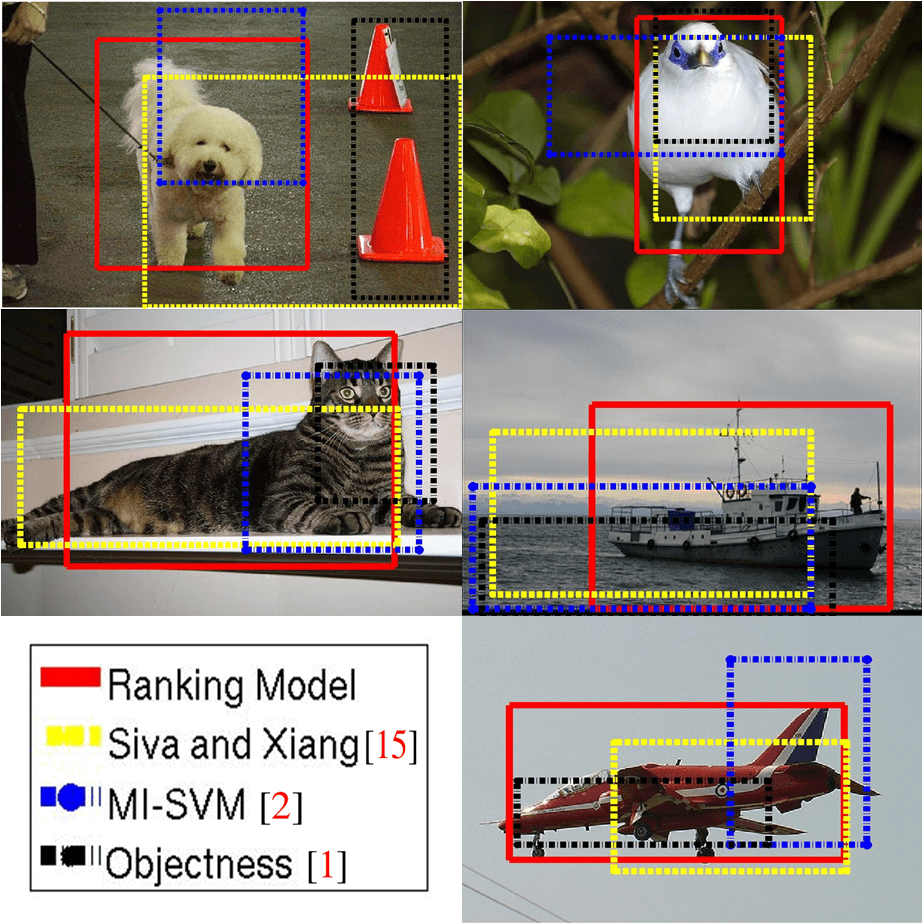}}
\caption{Qualitative results of our ranking model and Competitors}
\label{fig:figure1}
\end{minipage}
\hspace{0.7cm}
\begin{minipage}[b]{5cm}
\centering
\fbox{\includegraphics[height=6.6cm]{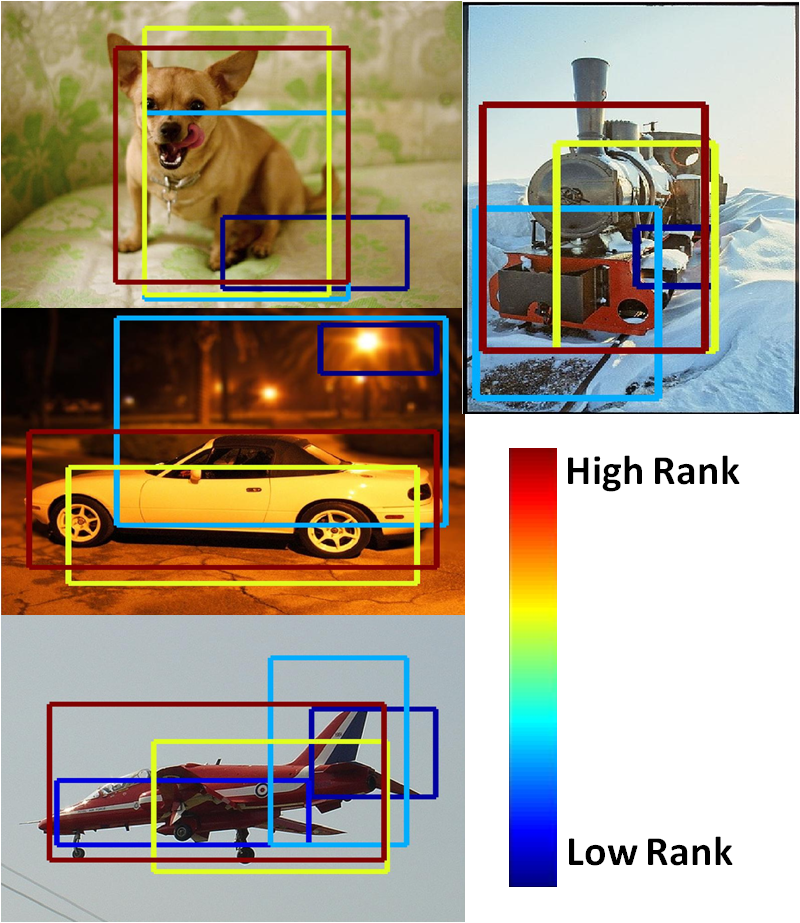}}
\caption{Examples of the ranking of salient regions by our model.}
\label{fig:figure2}
\end{minipage}
\end{figure}

\newcommand{\tabincell}[2]{\begin{tabular}{@{}#1@{}}#2\end{tabular}}
\begin{table}
\scriptsize
\begin{center}
\begin{tabular}{|c| c|c| c|}
\hline
\textbf{Class} & \textbf{Ranking Model}  & \textbf{Siva and Xiang} \cite{Sivaiccv2011} & \textbf{MI-SVM} \cite{Andrews03supportvector}\\
\hline
aeroplane & \textbf{54.74\%}  &  45.40 \%  & 37.80 \% \\
bicycle & \textbf{22.65 \%} &  20.60 \%  & 17.70 \% \\
bird & \textbf{33.71\%} &   29.70 \%  & 26.70 \% \\
boat & \textbf{24.45\%} &   12.20 \%  & 13.80 \% \\
bottle & 4.62\% &  4.10 \%  &  \textbf{4.90 \%} \\
bus & 33.90\% &   \textbf{37.10 \%}  & 34.40 \% \\
car & \textbf{42.48\%} &   41 \%  & 33.70 \% \\
cat & \textbf{57.04 \%} &   53.40 \%  & 46.60 \% \\
chair & \textbf{7.30\%} &  6.5 \%  & 5.4 \% \\
cow & \textbf{39.05\%} &  31.90 \%  & 29.80 \% \\
diningtable & \textbf{24.13\%} &  20.5 \%  & 14.5 \% \\
dog & \textbf{43.32\%} &  40.9 \%  & 32.80 \% \\
horse & \textbf{41.30\%} &  37.3 \%  & 34.80 \% \\
motorbike & \textbf{51.49\%} &  46.50 \%  & 41.60 \% \\
person & \textbf{25.34\%} &  22.3 \%  & 19.9 \% \\
pottedplant & \textbf{13.26\%} &   10.2 \%  & 11.4 \% \\
sheep & \textbf{27.97\%} &   27.1 \%  & 25 \% \\
sofa & 29.51\% & \textbf{ 32.30 \%}  & 23.60 \% \\
train &\textbf{ 54.55\%} &   49.00 \%  & 45.20 \% \\
tvmonitor & \textbf{11.76\%} &  9.8 \%  & 8.6 \% \\
\hline

\end{tabular}
\end{center}

\caption{Detailed annotation results for \emph{AllView} dataset}
\label{results_class_ranking}
\end{table}
\noindent \textbf{Ranking model vs. state-of-the-art WSL approaches --} We compare our ranking model with Objectness \citep{Alexewhatisobject}, \misvm~\cite{Andrews03supportvector} and the method of Siva and Xiang \cite{Sivaiccv2011}. Table \ref{results_ranking} shows that our ranking model can outperform all our competitors. Anecdotal results are shown in Fig.~\ref{fig:figure1}.
Objectness \cite{Alexewhatisobject} is a generic object detector which provides bounding boxes of generic foreground objects with a score indicating its ``objectness'' or the degree to which it is a foreground region. The region with the highest ``objectness'' is selected as the annotation. Our ranking model, as well as \misvm~\citep{Andrews03supportvector} and  Siva and Xiang \cite{Sivaiccv2011}, re-weights the top $N=100$ boxes chosen by the \cite{Alexewhatisobject} and as such is closely related to objectness. From Table~\ref{results_ranking} we can see that our ranking model outperforms the objectness measure by 8\% because it uses the auxiliary dataset to learn the salient region ranking that best overlaps with the ground truth. \misvm~ \cite{Andrews03supportvector}, like our method, starts by approximating the positive instance of each image as the average of the top $N=100$ instances then iteratively trains a SVM and re-annotates the target images. \misvm~ only uses weak annotation information from the target classes compared to our method which uses a strongly annotated auxiliary dataset. The superior performance of our model suggests that transferrable information learned on the strongly annotated dataset is more useful even though it contains object classes unrelated to the target class.
The most recent work on weakly supervised annotation is the method of Siva and Xiang \cite{Sivaiccv2011}. Their method fuses results from inter-class information and intra-class and saliency information to select the best annotation for each target image. It achieves the best annotation accuracy to date on the VOC 2007 data. We show that our ranking model, without using the inter-class and intra-class information, can achieve nearly 3\% improvement over the fused results. Again this shows the effectiveness of our model in extracting transferrable information from strongly annotated auxiliary data.

\noindent \textbf{Further evaluations on our method--} Table \ref{results_class_ranking} show the annotation accuracy of our method on each of the 20 classes. It gives some insight into where exactly the strength of our model lies. It can be seen that our method wins all but three classes but the performance is particularly strong on the boat (almost double the accuracy of the state-of-the-art methods) and aeroplane classes. Both classes have a very consistent background or strong contextual information: boats mostly appear on water and aeroplanes often appear in the sky. This pose a stern challenge to the existing MIL based approaches. Specifically, methods using intra-class information will tend to select water or sky which will appear more similar than the actual boats or aeroplanes that have a higher within class variance. Methods using inter-class information will struggle to eliminate water and sky (see Fig.~\ref{fig:figure1}) because they do not occur a lot in other classes. Inter-class methods will also have trouble with regions containing parts of the object and parts of the background. Unlike intra-class and inter-class methods, our method explicitly tries to rank the non-overlapped and partially overlapped regions lower than the highly overlapped regions (the aeroplane examples in Fig.~\ref{fig:figure1} and Fig.~\ref{fig:figure2} show clearly that those partially overlapped regions, selected wrongly by the alternative methods, are ranked lower by our model). This explains the superior performance of our method on these classes.

Our annotation accuracy of $32.13\%$ was obtained by approximating the ground truth feature using the average of the salient bounding box features for both the auxiliary and target datasets (Eqs.~\ref{eq:tapp} and \ref{eq:auxapp}). Alternatively, one could use the actual ground truth feature for training (Eq.~\ref{eq:diffgt}). This gives a slight drop in accuracy to $31.93\%$. This suggests that to make the learned model generalise better to  target classes, using the same approximation for the auxiliary data, even though it is not necessary, can be beneficial. We also notice that the annotation accuracy of our method varies little ($1.03\%$ standard deviation) across different trials. This verifies our assumption that the information we intend to transfer is not affected by the relationship between auxiliary and target object classes.

\begin{table}[ht]
\begin{center}
\begin{tabular}{l| c| c}
\hline
\textbf{Method} & \textbf{ByItself} & + \textbf{Ranking Model}\\
\hline
Objectness \cite{Alexewhatisobject} & 24.12\% &  \textbf{32.60\%}\\
\hline
MI-SVM \cite{Andrews03supportvector} & 25.12\% &  \textbf{31.23\%}\\
\hline
Siva and Xiang \cite{Sivaiccv2011} & 29.29\% &  \textbf{30.63\%} \\
\hline
\end{tabular}
\label{results_fusion_ranking}
\end{center}

\caption{Results of our ranking model fused with existing methods on \emph{AllView}}
\label{results_fusion}
\end{table}
\noindent \textbf{Fusion with existing methods --}
Since the information we use is very different from that used by existing methods, we combine our ranking model with existing methods including objectness \citep{Alexewhatisobject}, {\sc mi-svm} \cite{Andrews03supportvector} and \cite{Sivaiccv2011} by a score level fusion. The results are shown in Table \ref{results_fusion_ranking}. We find that combining our ranking model with existing methods consistently improves the existing method's performance. However, the best overall performance is obtained by combining our ranking model with the objectness score of \citep{Alexewhatisobject}. Ranking model plus objectness perform the best because they use related but complimentary cues -- objectness predicts the saliency based on cues such as image frequency and colour contrast, whereas our ranking model learns how the salient regions tend to overlap with the actual ground truth.

\begin{table}[ht]
\begin{center}
\begin{tabular}{l c c}
\hline
\textbf{Method} & \textbf{\emph{AllView} 20 Class} \\
\hline
Ranking Model & \textbf{32.13\%} \\
Generic Object Detector & 7.42 \%  \\
\hline
\hline
2Rank Model  & 30.89\% \\
Non-Ranking Model  & 20.45\% \\
\hline
\end{tabular}
\end{center}

\caption{Results of the alternative transfer methods}
\label{results_alternative}
\end{table}
\noindent\textbf{Alternative transfer learning methods --}
The strongly annotated auxiliary dataset can be used to learn alternative transfer learning models. Firstly, we consider learning a \textit{generic object detector}. All salient regions in the auxiliary dataset with greater than 50\% overlap with the ground truth are put in the positive class and the rest  in the negative class.  A linear SVM\footnote[1]{\scriptsize The SVM with non-linear kernel are too expensive to run on this dataset. Note that the ranking SVM implementation we adopted also uses linear kernel.} is then trained as a classifier on the raw \bow histograms $x_{i,j}$. The performance of the \textit{generic object detector} is reported in Table~\ref{results_alternative}. The generic object category will have a huge intra-class variance as well as be multi-modal in nature; importantly they are visually very different from the target object classes.  This basic transfer learning approach thus performs poorly.

\begin{table}[ht]
\begin{center}
\begin{tabular}{l c c}
\hline
\textbf{Method} & \textbf{\emph{SingleView} 6$\times$2 Class} \\
\hline
Ranking Model + Objectness & 39.74\% \\
Siva and Xiang (Fused) \cite{Sivaiccv2011}& 39.60\%  \\
\hline
Deselares \etal \cite{Deselaerslocalizing2010} & 35\% \\
\hline
Pandey and Lazebnik \cite{Pandeyiccv2011} \\
a. Before cropping  & 36.72\%\\
b. After cropping$^{*}$  & \textbf{43.73\%} \\
\hline
\multicolumn{2}{l}{{\scriptsize$^{*}$The cropping method can be applied to any other method to improve performance.}}
\end{tabular}
\end{center}

\caption{Annotation results for \emph{SingleView} dataset}
\label{results_singleview}
\end{table}

Second, we compare our ranking method with a non-ranking transfer learning to show the importance of formulating this as a ranking problem. For easier comparison, we consider the same setup as our ranking model but limit the number of ranks to two. That is, instead of using $N=100$ ranks, all salient regions are ranked either 1 or 2 during training depending on whether they are over 50\% overlap with ground truth. This model (named 2Rank model to distinguish it from our full ranking model) is compared with a non-ranking model which is essentially a binary linear SVM where the two ranks are used as the class label. Both models use the identical absolute difference features as input and differ only in the formulation of their cost functions. Table~\ref{results_alternative} shows that the ranking formulation leads to significantly better performance.

\noindent \textbf{Comparison to methods using single pose data --}
As a last comparison we evaluate the performance of our ranking model on the much simpler and smaller \textit{SingleView} dataset used in \cite{Deselaerslocalizing2010, Pandeyiccv2011, Sivaiccv2011}. \textit{SingleView} is a subset of \vocs~ that consists of six classes (aeroplane, bicycle, boat, bus, horse, and motorbike) where the left and right pose are considered as separate classes for a total of 12 classes, meaning both the presence of the object as well as the object pose needs to be manually annotated. For \textit{SingleView} we use the classes bird, car, cat, cow, dog and sheep as the auxiliary dataset per \cite{Deselaerslocalizing2010}. The results are presented in Table~\ref{results_singleview}. Our method is comparable to the competitors on this much easier task.

\section{Conclusion}

In this paper we presented a novel ranking based transfer learning method for object annotation, which effectively transfers a model for automatic annotation of object location from an auxiliary dataset to a target dataset with completely unrelated object categories. Our experiments demonstrate that our transfer learning based approach achieves higher accuracy than the state-of-the-art weakly supervised approaches.

\bibliography{egbib_zhiyuan}
\end{document}